\documentclass[]{interact}

\usepackage{epstopdf}
\usepackage[caption=false]{subfig}

\usepackage{url}
\usepackage{hyperref}           

\usepackage[utf8]{inputenc} 
\usepackage[natbibapa,nodoi]{apacite}
\graphicspath{{/home/giorgio/Dropbox/My_Papers/Articolo_JORS_01_20/Articolo_Latex/Revision/images/}{C:/Users/eo7giovisani/Desktop/Dropbox/My_Papers/Articolo_JORS_01_20/Articolo_Latex/Revision/images/}}

\theoremstyle{plain}

\theoremstyle{definition}

\theoremstyle{remark}

\usepackage[ruled,vlined,linesnumbered]{algorithm2e}
\newcommand{\nonl}{\renewcommand{\nl}{\let\nl\oldnl}}

\usepackage[noend]{algorithmic}
\usepackage{xspace}
\newcommand{\ovl}{\textsc{overlap}\xspace}
\newcommand{\CI}{\textsc{ConfInt}\xspace}

\newcommand{\feat}{\texttt{feat}\xspace}
\newcommand{\pair}{\texttt{pair}\xspace}
\newcommand{\CIpair}{\texttt{CIpair}\xspace}
\newcommand{\conc}{\textsc{concordance}\xspace}
\newcommand{\combVSI}{C_2^m(g_1...g_m)\xspace}
\newcommand{\combCSI}{C^{| \mathcal{M}|}_2\xspace}
\newcommand{\VSI}{\textsc{VSI}\xspace}
\newcommand{\CSI}{\textsc{CSI}\xspace}

\usepackage{amssymb}
\let\emptyset\varnothing
\usepackage[official]{eurosym}

\begin{document}


\title{Statistical stability indices for LIME: obtaining  reliable explanations for Machine Learning models}

\author{
\name{Giorgio Visani\textsuperscript{a,b}\thanks{CONTACT Giorgio Visani. Email: giorgio.visani2@unibo.it}, Enrico Bagli\textsuperscript{b}, Federico Chesani\textsuperscript{a}, Alessandro Poluzzi\textsuperscript{b} and Davide Capuzzo\textsuperscript{b}}
\affil{\textsuperscript{a}Università degli Studi di Bologna, Dipartimento di Ingegneria e Scienze Informatiche, viale Risorgimento 2, 40136 Bologna (BO), Italy; \\
\textsuperscript{b}CRIF S.p.A., via Mario Fantin 1-3, 40131 Bologna (BO), Italy}
}

\maketitle

\begin{abstract}
Nowadays we are witnessing a transformation of the business processes towards a more computation driven approach. The ever increasing usage of Machine Learning techniques is the clearest example of such trend.
This sort of revolution is often providing advantages, such as an increase in prediction accuracy and a reduced time to obtain the results.
However, these methods present a major drawback: it is very difficult to understand on what grounds the algorithm took the decision.
To address this issue we consider the LIME method.
We give a general background on LIME then, we focus on the stability issue: employing the method repeated times, under the same conditions, may yield to different explanations.
Two complementary indices are proposed, to measure LIME stability.
It is important for the practitioner to be aware of the issue, as well as to have a tool for spotting it. Stability guarantees LIME explanations to be reliable therefore a stability assessment, made through the proposed indices, is crucial.
As a case study, we apply both Machine Learning and classical statistical techniques to Credit Risk data. We test LIME on the Machine Learning algorithm and check its stability. Eventually, we examine the goodness of the explanations returned.
\end{abstract}

\begin{keywords}
Credit Scoring; Machine Learning; Explainability; LIME; Stability
\end{keywords}

\section{Introduction}

Nowadays, more and more interest is devoted to the concept of "learning from the data", i.e. using the data collected about the process to predict its outcome \citep{hastie_elements_2009}.
The main ingredients of its recent success are the huge availability of data sources and the increased computational power, which allows complex algorithms to deliver results in a relatively short time.

\medskip

In statistics, making predictions about the future is a particularly relevant topic. To address the subject, simple algorithms and methods have been developed over the years, the most famous being Linear Regression and Generalised Linear Models \citep{greene_econometric_2003}. However, with the advent of powerful computing tools, more sophisticated techniques have been developed.
In particular, Machine Learning models are able to perform intelligent tasks usually done by humans, supporting the automation of data driven processes.
\medskip

Despite the enhanced accuracy, Machine Learning models display weakness especially when it comes to interpretability, i.e. “the ability to explain or to present the results, in understandable terms, to a human" \citep{hall_introduction_2018}.
They usually adopt large model structures and refine the prediction using a huge number of iterations. The logic underlying the model ends up hidden under potentially many strata of mathematical calculations, as well as scattered across a too vast architecture, preventing humans from grasping it.
\medskip


To achieve the interpretability, quite a few techniques have been proposed in recent literature. These approaches can be grouped based on different criteria \cite{molnar_interpretable_2020}, \cite{guidotti_survey_2018} such as i) Model
agnostic or model specific ii) Local, global or example based iii) Intrinsic or post-hoc iv) Perturbation or saliency based.
Herein, we focus on LIME (Local Interpretable Model-agnostic Explanations), a local interpretability framework, developed by \citealp{ribeiro_why_2016}.
\medskip

The technique may suffer from a lack of stability, namely repeated applications of the method  under the same conditions may obtain different results. 
This is a particularly delicate issue however it is rarely taken into consideration. Even worse, many times the issue is not spotted at all, e.g. when just a single call to the method is done and the result is considered to be okay without further checks.
\medskip

In this paper, we introduce a pair of complementary stability indices, useful to measure LIME stability and spot potential issues. They represent an innovative contribution to the scientific community, addressing an important research question.
\par
The indices are calculated on repeated calls of the method, to evaluate the similarity of the results. They may be applied on every trained LIME method and will allow the practitioner to be aware about potential instability of the results, otherwise to ensure that the trained method is consistent.
\medskip

Hereafter, a brief introduction on the explainability techniques is presented in Chapter 2. The LIME technique is exhaustively analysed in Chapter 4, including its weak points. A thorough discussion about LIME stability can be found in Chapter 4, along with a description of some recent works tackling the issue. Our proposition is extensively discussed in Chapter 5.
Eventually, a practical application of the method in the Credit Risk Modelling field is shown in Chapter 6.
Chapter 7 is dedicated to Discussion and Conclusions.
\medskip

The code used for the experiments is available at\\ \href{https://github.com/giorgiovisani/LIME_stability}{https://github.com/giorgiovisani/LIME\_stability}.

\section{Related Work}

Explainable methods are grouped into Global and Local Explainability techniques \citep{guidotti_survey_2018}. Global methods aim to give an understanding of the model as a whole: the explanation should apply to all the records in the dataset. Local methods instead, attempt to provide very good understanding just for a small portion of records.
\par
In the following review we consider both global and local techniques developed in a model agnostic fashion, so as to be effective on any kind of ML model by construction.
\medskip

A popular approach is to exclude a certain feature, or group of features, from the model and evaluate the loss incurred in terms of model goodness. The idea has been first introduced by Breiman \citep{breiman_random_2001} for the Random Forest model and has been generalised to a model-agnostic framework, named LOCO \citep{lei_distribution-free_2018}. Based on variable exclusion, the predictive power of the ML models has been decomposed into single variables contribution in PDP \citep{friedman_greedy_2001}, ICE \citep{goldstein_peeking_2015} and ALE \citep{apley_visualizing_2016} plots, based on different assumptions about the ML model. The same idea is exploited also for local explanations in SHAP \citep{lundberg_unified_2017-1}, where the decomposition is obtained through a game-based setting.\\
These methods' goal is a fair measure of feature importance. They usually suffer correlation among features since it introduces distortion in the results, while the changes proposed to tackle the correlation have stripped the techniques of some theoretical properties.
\par
Another common approach is to train a surrogate model mimicking the behaviour of the ML model. In this vein, approximations on the entire input space are provided in \citep{craven_extracting_1996} and \citep{zhou_interpreting_2016-1} among others, while LIME \citep{ribeiro_why_2016} and its extension using decision rules \citep{ribeiro_anchors_2018-1} rely on this technique for providing local approximations.\\
Surrogate models have the nice perk of exploiting a prediction model, this allows to make some sort of what-if analysis (\textit{\small eg. If I were to earn \euro{}5000 more a year,  how many points would I gain on my credit score?}), which is not possible for feature attribution methods. Although one should pay attention to their limitations: global techniques are usually a coarse estimate of the ML model, so the what-if analysis can be quite approximate; on the contrary local methods provide good approximation but just for a small region of the input variables, this means the scenario we test should comprise just small changes.

\section{LIME}

LIME \citep{ribeiro_why_2016} is a method for explaining black-box models, i.e. models whose inner logic is hidden and not clearly understandable. \\
It provides a number of explainable models which closely resemble the original model behaviour. Each model is specific for an input point $x$: only in its neighbourhood the  explainable model's predictions are guaranteed to be very close to the black-box ones. This peculiarity places LIME among the Local Explainability tools.\\
In the following we will focus specifically on LIME for tabular data, since they represent the vast majority of data sources in the Credit Scoring field.
\medskip

\subsection{General Idea}
LIME aims to approximate the black-box model $f$ with a simple function $g$ around the point of interest $x$. $g$ is required to lie into the class of explainable models $G$. 

\begin{align*}
    \begin{aligned}
    &f :\mathbb{R}^P \rightarrow \mathbb{R}, &\text{black-box model} \\
    &g : \mathbb{R}^p \rightarrow  \mathbb{R}, &\text{explainable model}
    \end{aligned}
\end{align*}

where $P$ is the number of features employed by the black-box model, to make predictions about the response variable. The explainable model $g$ uses only $p$ of the original $P$ variables, in order to reduce the complexity. \\
Solving the following optimisation problem, we obtain the function $g$ most similar to $f$ in the neighbourhood of $x$.

\begin{align*}
    \begin{aligned}
        arg \min \limits_{g \in G} &\;L(f,g,\pi_x)+\Omega(g) \\
         \Omega(g):&\; \text{complexity of g}\\
         L:&\; \text{loss function}\\
         \pi_x:&\; \text{weight assigned according to $x$ proximity}
    \end{aligned}
\end{align*}

\begin{figure}[h]
\centering
\includegraphics[width=0.5\textwidth]{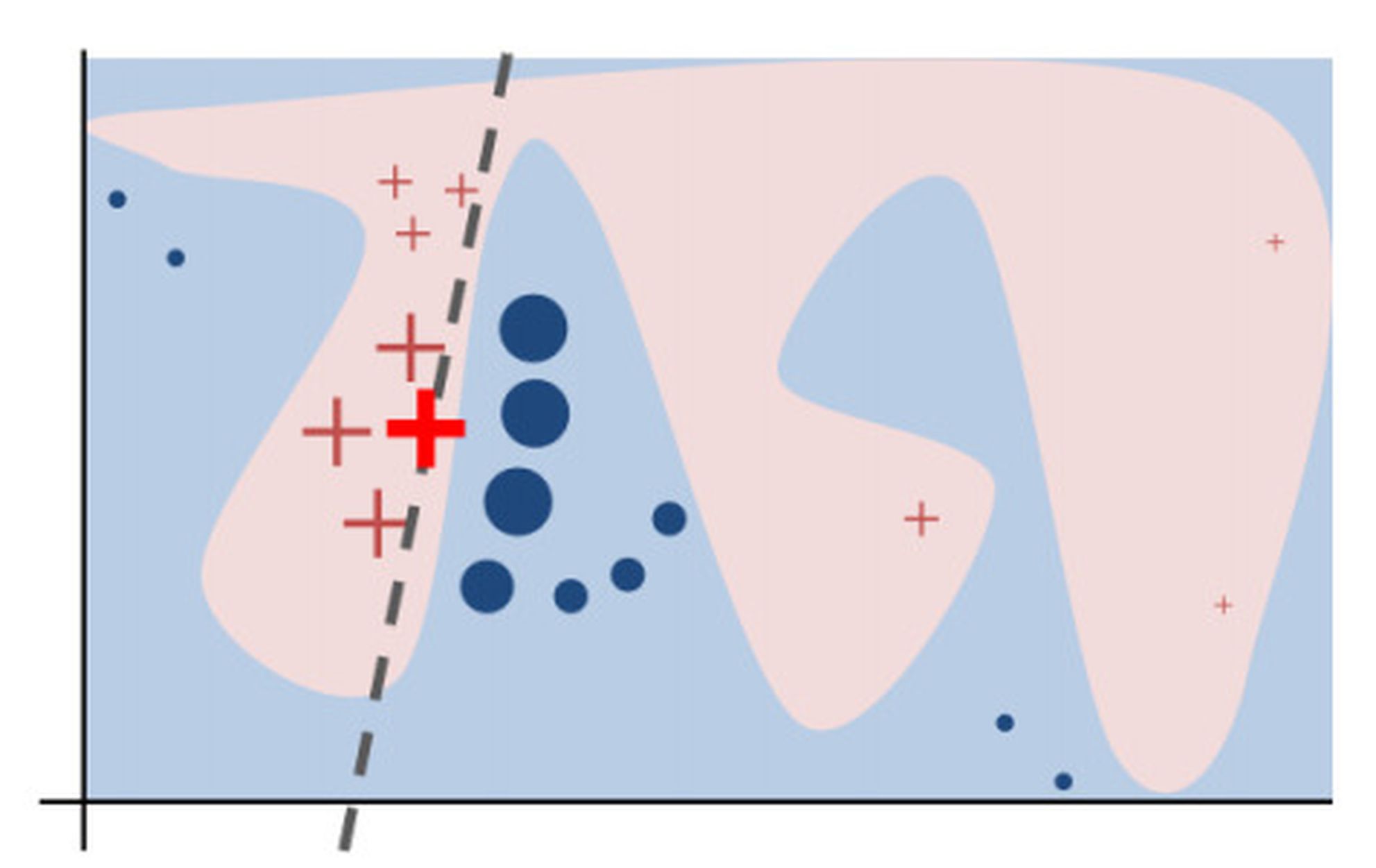}
\captionsetup{format=hang}
\caption{LIME’s modus operandi. \\
Courtesy of \citealp{ribeiro_why_2016}}
\label{LIME}
\end{figure}

Chosen a given individual $x$, LIME returns a local explainable model $g$, which in turn provides the most important variables to predict the points in the $x$ neighbourhood (see Figure \ref{LIME}).\\

\subsection{LIME Algorithm in detail}

LIME relies on producing new points, generated from a multivariate distribution of the features in the dataset. The features are considered to follow a Normal distribution, whose parameters are inferred from the dataset. For the purpose of data generation, each feature is assumed to be independent from the others. \\
The points are generated all over the space of the dataset variables.
\smallskip

In order to account for locality, LIME weights each new point using a Gaussian Kernel. Its purpose is to assign a weight $\pi_x$ to each point, based on its distance from the individual to be explained.
\smallskip

Next step is to query the black-box model and obtain the predicted values for the new points.
Doing so, we end up with a brand new dataset.
\smallskip

Such dataset undergoes a rescaling process, which standardises each $X$ feature, leaving untouched the response variable. This allows to compare the contribution of each feature on a similar scale, namely the standard deviation of each variable.
\par 
In order to obtain a human-understandable linear model, it is mandatory to use only a bunch of features. Therefore LIME performs also a feature selection step, done usually with Lasso technique. This allows the explanations to be compact and human readable. The number of variables to be retained, namely $p$, is decided by the practitioner.  
\par 
On the standardised $p$-dimensional dataset, LIME performs Ridge Regression, i.e. a Linear Regression combined with a penalty related to the $\ell 2$ norm of the coefficients \citep{hoerl_ridge_1970}, used to prevent overfitting. The model training is done in a weighted fashion: each point contributes to the model according to its weight $\pi_x$.
\smallskip

The result is a linear model, which provides understanding of the process through its coefficients: the higher the coefficient, the bigger the variation in the value of the response variable when the feature is changed. The sign of the coefficient tells us the direction of the variation, namely if we will face a decrease or an increase of the output value.

\subsection{LIME Drawbacks}

LIME is sensitive to the dataset dimensionality: when it is employed to interpret a Machine Learning model built using a huge number of variables, the local explanation is unable to discriminate among relevant and irrelevant features.
\smallskip

This phenomenon is due to the weighting kernel. Generally speaking, it can be considered as a similarity (or distance) function, thus it inherits the drawbacks of this class. As thoroughly described by \citealp{beyer_when_1999}, in high dimensional datasets, chosen a fixed point, the distance to its nearest data point approaches the distance to the farthest one, as dimensionality increases. \\
LIME applies the kernel function before variable reduction, thus for high dimensional datasets, the kernel is not able to distinguish between near and distant points, considering all of them approximately at the same distance. This results in a loss of the locality concept and consequently in a bad performance of the algorithm.
\smallskip

\begin{figure}[h]
\centering
\includegraphics[width=0.7\textwidth]{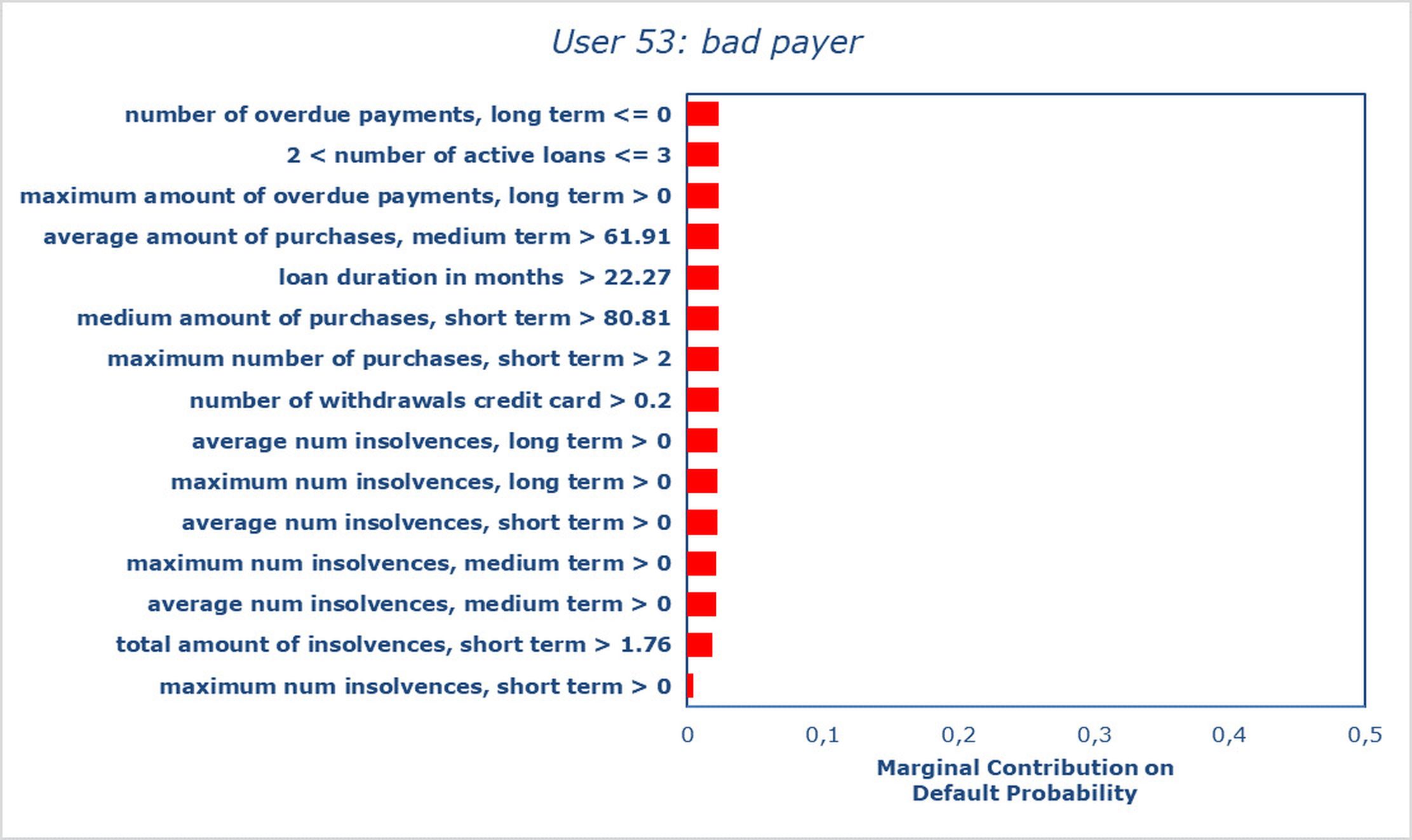}
\caption{LIME explanations are not informative
when applied to Machine Learning
models with many
independent variables, in this case Gradient Boosting model using 100 features.}
\label{LIME_100_var}
\end{figure}

Such occurrence is intuitively shown in Figure \ref{LIME_100_var}. In it, the Credit Scoring dataset used in Section \ref{Credit_Scoring_use_case} has been employed to train a Gradient Boosting Tree model using 100 variables. Although the Gradient Boosting has shown good performance in such a setting, LIME applied to the model has not been able to discriminate among important and irrelevant regressors. In particular, many features exhibit low values and almost all of them are equally important.
\medskip

This weakness curbs LIME's employment on black-box models handling high dimensional datasets. To date, it is a practitioner duty to ensure the dataset dimensionality is low enough for LIME to work well. This usually requires feature selection, upstream of data modelling.

\section{LIME Stability issue}

Consider choosing a specific individual and performing LIME on it, several times. Indeed, it is desirable to obtain the same explanations from each call.
\par 
Every time LIME is employed, it generates new data points, which follow the same distribution (law) but are different among distinct applications. This is due to the random nature of the sampling.
Using different points it may happen to obtain divergent explainable models $g$, thus different explanations, for the chosen individual.
\par 
Based on this evidence, we define the concept of LIME stability: explanations derived from repeated LIME calls, under the same conditions, are considered stable when statistically equal.
\medskip

In \cite{alvarez-melis_robustness_2018-1} the authors provide insight about LIME's lack of robustness, a similar notion to the above-mentioned stability. Analogous findings also in \cite{gosiewska_ibreakdown_2019}.
\par

Some approaches, grouped in two high level concepts, have been recently laid out in order to solve the stability issue.

\subsubsection*{Avoid the sampling step}

In \cite{zafar_dlime_2019} the authors propose to bypass the sampling step using the training units only and a combination of Hierarchical Clustering and K-Nearest Neighbour techniques. Although this method achieves stability, it may find a bad approximation of the ML function, in regions with only few training points. 

\subsubsection*{Evaluate the post-hoc stability}

The shared idea is to repeat LIME method at the same conditions, and test whether the results are equivalent. Among the various propositions on how to conduct the test, in \cite{shankaranarayana_alime_2019} the authors compare the standard deviations of the Ridge coefficients, whereas \cite{molnar_limitations_2020} examines the stability of the feature selection step - whether the selected variables are the same - .
\medskip

Although we consider recent work on the topic headed in the right direction, we feel more work has to be done in order to provide solid grounds and mathematical rigour to the metrics evaluating LIME stability.  

\section{Our Proposition}

Hereafter, a formal description of the framework considered, in order to evaluate LIME stability. \\
Consider the black-box model $f$ composed by $P$ variables, a chosen point to be explained $x$ and a fixed number of variables $p$, used in LIME's explainable models.
The Weighted Ridge Regression model $g$ (LIME's output) can be viewed as a mapping function between the set of variables and the respective coefficients.

\begin{equation}\label{g_explainable_model}
    g : \mathcal{F} \rightarrow \mathbb{R}
\end{equation}
where $\mathcal{F}$ is the set of variables, of cardinality $P$. $p$ out of $P$ features will be associated with a value different from 0, the others $P-p$ variables will have 0 coefficient, meaning they are irrelevant to the model. The formulation $g(\feat)$ indicates the coefficient value of the feature named \feat, in the model $g$.
\medskip

We perform $m$ different calls to LIME on the model $f$ and the individual $x$, obtaining $m$ different explainable models $g_1 ... g_m$.
\medskip

We want to: (\textit{i}) check whether different $g$ are composed by the same variables, (\textit{ii}) compare the coefficients of the same variable among $g_1 ... g_m$ and test whether they can be considered equal.
\medskip

To this purpose, we devise two complementary indices: the Variables Stability Index (\VSI) and Coefficients Stability Index (\CSI).

\subsection{Variables Stability Index: \VSI}
The Variables Stability Index (\VSI), whose steps are explained in Algorithm \ref{VSI_Algortihm}, addresses the first point, namely it compares the variables composition of the $g_1 ... g_m$ models.
\medskip

We consider the set $C_2^{m}(g_1 ... g_m)$ of all possible combinations of the $m$ explainable models, two by two. The generic element of $C_2^{m}(g_1 ... g_m)$ is the pair $(g_\alpha,g_\beta)$.
\medskip

We define a measure of concordance among the two explainable models in each pair:
\begin{align}
    \begin{aligned}
    &\pair = (g_\alpha,g_\beta)\\
    &\mathcal{F}_\alpha = \{ \feat \in \mathcal{F} : g_\alpha(\feat) \neq 0 \} \\
    &\mathcal{F}_\beta = \{ \feat \in \mathcal{F} : g_\beta(\feat) \neq 0 \} \\
    &\conc(\pair) = \left| \mathcal{F}_\alpha \cap \mathcal{F}_\beta \right|
    \end{aligned}
\end{align}

where $\mathcal{F}_\alpha$ and $\mathcal{F}_\beta$ represent respectively the variables used in the explainable models $g_\alpha$ and $g_\beta$. The \conc function returns an integer value, namely the cardinality of the intersection between $\mathcal{F}_\alpha$ and $\mathcal{F}_\beta$, ranging from 0 to $p$. It represents the number of variables used by both $g_\alpha$ and $g_\beta$.
\medskip

\begin{algorithm}[h]\label{VSI_Algortihm}
\SetAlgoLined

\KwIn{$g_1 ... g_m$}
$n = 0$ \\
\For{$\pair \text{ in } \combVSI$}{
$n = n + \frac{\conc(\pair)}{p}$
}
$\VSI = \frac{n}{\left| \combVSI \right|}$\\
\KwOut{\VSI}
 \caption{Variables Stability Index (\VSI)}
\end{algorithm}
\medskip

We evaluate the \conc over all the pairs in $\combVSI$ and we average them, obtaining the \VSI index, ranging from 0 to 1. We express the index as a percentage: it now spans from 0 to 100, the more it approaches 100 the more the variables found in different applications are the same.
\medskip

\subsection{Coefficients Stability Index: \CSI}
The equality between coefficients of the different $g_1 ... g_m$ models is now under investigation. In the following, we derive the statistical distribution of the coefficients and we rely on it, to create confidence intervals and possibly statistical tests.
\medskip

It is a well-known result \citep{greene_econometric_2003}, that under the classic assumptions of Linear Regression, the coefficients are guaranteed to follow a Gaussian distribution. This is not sufficient, since we deal with Weighted Ridge Regression.
\medskip

In \cite{van_wieringen_lecture_2015}, the distribution of the Ridge Regression estimator is given by the formula: 
\begin{equation}\label{Ridge_law}
        \hat{\beta}(\lambda) \sim \mathcal{N}\Big( (X^TX+\lambda I_{p})^{-1} X^TX\beta,\sigma^2 (X^TX+\lambda I_{p})^{-1} X^TX [(X^TX+\lambda I_{p})^{-1}]^T \Big)
\end{equation}

where $X$ is the matrix of observations. In our setting, $X$ is composed by the points randomly sampled inside LIME. The matrix $I_p$ stands for the identity matrix (dimensions $p \times p$). $\sigma^2$ is the variance of the $\mathcal{E}_i$ random variables describing the errors per each sampled point. Under the Regression assumptions the errors $\mathcal{E}_i$ are independent and identically distributed (IID) following a Gaussian law: $\mathcal{E}_i \sim \mathcal{N}(0, \sigma^2)$. $\lambda$ stands for the Ridge regularisation coefficient.
The vector $\beta$ represents the true values of the coefficients in population, whereas $\hat{\beta}$ consists in the estimates of the true values, using the $X$ dataset.
\smallskip

In our setting, we may consider the $\beta$ values as the unknown coefficients of the best linear approximation of $f$ in the neighbourhood of $x$. LIME aims to provide $\hat{\beta}$ closest as much as possible to the unknown $\beta$ values.
\medskip

Concerning Weighted Regression, it is usually estimated via Generalised Least Squares (GLS) which guarantee the distribution of its estimators to be the following \citep[see][]{johnston_econometric_1972} :
\begin{equation}\label{Weighted_law}
    \hat{\beta} \sim \mathcal{N}\Big( (X^TWX)^{-1} X^TWX\beta,\sigma^2 (X^TWX)^{-1} \Big)
\end{equation}

In the formula, $W$ is the $n \times n$ diagonal matrix of weights per each unit. In our setting, the $W$ matrix is populated by the kernel weights calculated on the distance of each sampled point from $x$.
\medskip

It is important to recall that $\sigma^2$ is an unknown value and we are requested to obtain an unbiased estimator inferred from the data.
Such estimator takes the form: 
\begin{equation*}
    \hat{\sigma}^2 = \frac{\mathcal{E}W\mathcal{E}^T}{n-p}
\end{equation*}
for the Weighted Regression, as stated in \citet{johnston_econometric_1972}. $\mathcal{E}$ stands for the vector of the errors per each sampled point: $\mathcal{E} = (\mathcal{E}_1...\mathcal{E}_n)$. As far as Ridge Regression is concerned, the variance estimator remains unchanged from the Linear Regression's one \citep{van_wieringen_lecture_2015}.
\medskip

Using the building blocks stated before, we derive the distribution of the Weighted Ridge Regression estimator.
Starting from the Ridge Regression law (Equation \ref{Ridge_law}), we know \citep{billingsley_probability_2008} that the Gaussian distribution is invariant whenever we employ a matrix of known weights. This guarantee the Weighted Ridge law of the coefficients to be Gaussian.
Its distribution is \footnote{We do not state all the derivations of the expectation and variance formulae, for the sake of readability.}: 
\begin{align} \label{Weighted_Ridge_law}
    \begin{aligned}
    \hat{\beta}(\lambda) \sim \mathcal{N}\Big( &(X^TWX+\lambda I_{p})^{-1} X^TWX\beta,\\
    &\sigma^2 (X^TWX+\lambda I_{p})^{-1} X^TWX [(X^TWX+\lambda I_{p})^{-1}]^T \Big)
    \end{aligned}
\end{align}

We provide also the formula for the variance estimator of the Weighted Ridge Regression \footnote{In this formulation, we consider $\mathcal{E}$, as the errors of the Linear Regression model. In other words, the Weighted Ridge $\hat{\sigma}^2$ estimator is the same of Weighted Regression.\\
We may not use the errors of any Ridge model to calculate an unbiased estimator of the error variance $\sigma^2$, because Ridge regularisation term decreases the variance. Using such errors would cause the estimator to be biased towards 0.}:  
\begin{equation}\label{variance_estimator}
    \hat{\sigma}^2 = \frac{\mathcal{E}W\mathcal{E}^T}{n-p}
\end{equation}
where $n$ is the number of data points sampled inside LIME, $p$ denotes the number of variables considered in the explainable model.
\bigskip

Knowing the distribution of the coefficients, we might derive a test statistic to assess a null hypothesis of equality. This comparison can be carried out also among coefficients of two different regression models, as long as they were estimated on two independent samples drawn from the same law, as derived by \citealp{brame_testing_1998} for the coefficients of two distinct Linear Regressions.
\medskip

This assumption holds true in our experimental design, since the data are sampled from the features' distribution inferred from the original data. It means that the true generating distribution of $X_1 .. X_m$, i.e. the datasets sampled in repeated LIME calls, is identical, while the differences among them are attributable only to the sampling variance.
\par
Unfortunately, the simplifications carried out in \citealp{brame_testing_1998} and \citealp{greene_econometric_2003} in order to derive the t-test statistic, rely on the equality of the expected value of the two coefficients taken into consideration. This is true in Linear Regression, but the framework brakes down using a regularisation technique such as Ridge: the regulariser trades off the unbiasedness of the estimator in exchange for a possibly strong reduction of the variance.
\medskip

Since the estimator is not unbiased any more, the expected value is now depending on the design matrix $X$. As stated before, different LIME calls give rise to different design matrices, this implies the expected values of a specific variable, taken from two different explainable models $g_\alpha$ and $g_\beta$, to be different: $\mathrm{E}[g_\alpha(\feat) - g_\beta(\feat)] \neq 0$. This result causes the derivation of the t-test statistic to break down.
\bigskip

Testing the null hypothesis of equality has proven tricky and not easily solvable, hence we rely on the Gaussian distribution of the coefficients to construct 95\% confidence intervals. To do that, we design the function \CI, taking as input a $g$'s coefficient and giving back its confidence interval:
\begin{align}\label{ConfInt}
    \begin{aligned}
    \CI(g(\feat)) = [ &g(\feat) - 1.96 \;  \sqrt{\text{Var}(g(\feat))} \;, \\
    &g(\feat) + 1.96 \;  \sqrt{\text{Var}(g(\feat))} ]
    \end{aligned}
\end{align}

where $\text{Var}(g(\feat))$ is calculated based on the distribution given in Equation \ref{Weighted_Ridge_law}.
\medskip

We may consider the parameters to be different, within a 5\% error rate, when the confidence intervals are not overlapped at all. Instead, we consider them to be stable whenever the confidence intervals overlap to some extent.
\medskip

To this purpose, we devise the binary function \ovl, which takes as input a generic pair of confidence intervals $\CIpair = (\texttt{CI}_\alpha,\texttt{CI}_\beta)$ and returns either value 1 or 0, based on the overlap presence.

\begin{equation}
    \ovl(\CIpair) = \begin{cases}
    0 & \text{if } \texttt{CI}_\alpha \cup \texttt{CI}_\beta = \emptyset\\
    1 & \text{otherwise}
    \end{cases}
\end{equation}

The comparison among confidence intervals is carried out separately for each variable.
Chosen a certain feature, we check through the $g_1 ... g_m$ explainable models if the feature is relevant (coefficient different from 0). Whenever this happens, we build the confidence interval for the coefficient, using the function \CI, and we consider the set of all confidence intervals, namely $\mathcal{M}$, for the chosen variable.\\
We create all the possible combinations of the $\mathcal{M}$ items, two by two. This results in the set $\combCSI$, whose generic element is $\CIpair = (\texttt{CI}_\alpha,\texttt{CI}_\beta)$.
We calculate the overlap between the two intervals, using the \ovl function, for all the pairs in $\combCSI$.
\medskip

The outcome is a count variable, which we normalise dividing by the cardinality of the set $\combCSI$. The value obtained ranges from 0 to 1 and it is called the Partial Index (\textsc{Par}) for the variable considered. It represents a measure of concordance of the specific variable's coefficients among different LIME calls. \\
To achieve a general concordance metric, we average the Partial Indices of all the features and obtain the Coefficients Stability Index (\CSI), ranging from 0 to 1. Consider now the index as a percentage, rescaling it from 0 to 100: the more \CSI approaches 100, the more LIME coefficients may be considered stable in the neighbourhood of the chosen individual.
\medskip

\CSI steps are detailed in Algorithm \ref{CSI_Algortihm}.
\medskip

\begin{algorithm}[h]\label{CSI_Algortihm}
\SetAlgoLined

\KwIn{$g_1 ... g_m$}
\For{\feat in $\mathcal{F}$}{
$\mathcal{M} = \{\}$\\
\For{$i$ in $1 ... m$}{
\If{$g_i(\feat) \neq 0$}{
\texttt{CI} = \CI($g_i(\feat)$)\\
$\mathcal{M} = \mathcal{M} \; \cup$ \texttt{CI}
}
}

$n = 0$\\
\For{\CIpair in $\combCSI$}{
\If{\ovl(\CIpair)}{
$n ++$
}
}
$\textsc{Par}_\feat = \frac{n}{\left| \combCSI \right|}$ \\
}
\CSI = mean(\textsc{Par})\\
\KwOut{\CSI}
\caption{Coefficients Stability Index (\CSI)}
\end{algorithm}

\subsection{Interpretation of the indices}

The previously defined indices constitute a useful tool for assessing LIME stability in practical scenarios.
By construction, \VSI measures the concordance of the variables retrieved,  whereas \CSI tests the similarity among coefficients for the same variable, in repeated LIME calls.
\medskip

Both of them range from 0 to 100.\\
High \VSI values guarantee the variables retrieved in different LIME are almost always the same. On the contrary, low values testify explanations are not trustworthy: we may retrieve completely different variables explaining the same Machine Learning decision, according to different LIME calls.
\par
As far as \CSI is concerned, high values ensure LIME coefficient for each feature is reliable. Low values, instead, induce the practitioner to be very cautious: given a feature, the first LIME call will give back a certain value of the coefficient, but the one after is likely to retrieve a different value. Since the coefficient represents the impact of the feature on the Machine Learning decision, obtaining different values correspond to very different explanations.
\medskip

Each index has a proper meaning and checks for a particular stability instance. Achieving high values for both of them ensures stability, however low values for only one metric are still possible. Keeping the measurements separated allows for understanding which one of the two complementary definitions of stability has been violated by the trained LIME method.

\section{Practical Application to Credit Risk Data} \label{Credit_Scoring_use_case}

Credit Risk Modelling (CRM) consists in estimating the probability that a debtor will not repay the due amount. This task is regarded as a fully-fledged prediction task and as such, a variety of different learning techniques have been applied over the years in order to solve it.
\medskip

Since long, statistical approaches have been exploited, forming the core techniques in this field. Recently also Machine Learning models have been given a chance. They have usually shown an increase in the prediction power, although they do not provide reliable explanations for the scores they come up with.
\medskip

This is a particularly delicate issue in CRM, since it is a highly regulated field: GDPR \citep{kingston_using_2017}, as well as the "Ethical Guidelines for trustworthy AI" \citep{hleg_ethics_2019} and the Report from the "European Banking Authority" \citep{eba_report_2020}
testify the care dedicated to such topics by the European Community.
\medskip

In order to exploit the Machine Learning potential in the Credit Scoring field, it is mandatory to address the interpretability issue. To do that, our proposal concerns applying LIME on top of a well performing black-box algorithm. By doing so, we wish to retain the increased predictive power of Machine Learning, while providing meaningful explanations to the applicants involved, as well as to the regulator.
\medskip

To validate the above mentioned approach, we use a real-life dataset representative of a loan application process. It comes from an anonymised statistical sample, obtained by pooling data from several Italian financial institutions. \\
In it, there are several demographic, economic and financial variables used as predictors, whereas the response variable consists of only two categories (bad payer, good payer), framing the problem as classification. \\
The dataset composition is shown in the Table \ref{tab:1}. 
\medskip

\begin{table}[h]
\tbl{Dataset Composition \medskip}
{\begin{tabular}{|c|c|c|}
\cline{1-3}
\textbf{Data set name} & \textbf{Population} & \textbf{\%Bad} \\ \cline{1-3}
Train set           & 39.418              & 2,9\%  \\ \cline{1-3}
Test set               & 16.893              & 3,1\%  \\ \cline{1-3}
Total                  & 56.311              & 3\%   \\ \cline{1-3}
\end{tabular}}
\label{tab:1}
\end{table}
\medskip

Before using learning techniques of any kind, we take care of selecting the important features among the many variables available in the dataset. This is done for two reasons: (\textit{i}) classical models are not performing well in high dimensional settings and (\textit{ii}) LIME applied to high dimensional Machine Learning models would cause the method to fail. To this end, we retained only the most important 20 features, to be employed for learning purposes.

\subsection{Logistic Regression model}

On the dataset described above, we employ Logistic Regression along with a Machine Learning model, for the sake of comparison among the two techniques. \\
We consider Logistic Regression as the benchmark model, being explainable and widely used in Credit Scoring.
The technique is well described in \citealp{agresti_foundations_2015} and it usually obtains satisfactorily results on CRM datasets.
Logistic Regression provides also interpretability out-of-the-box: the parameters derived from the best curve’s estimation, can be regarded as odds ratio, i.e. the ratio between the probability of default and non-default, namely $\frac{P(Y=1|X=x)}{P(Y=0|X=x)}$.

\subsection{Machine Learning Model}

\begin{figure}[h!]
\centering
\includegraphics[width=\textwidth]{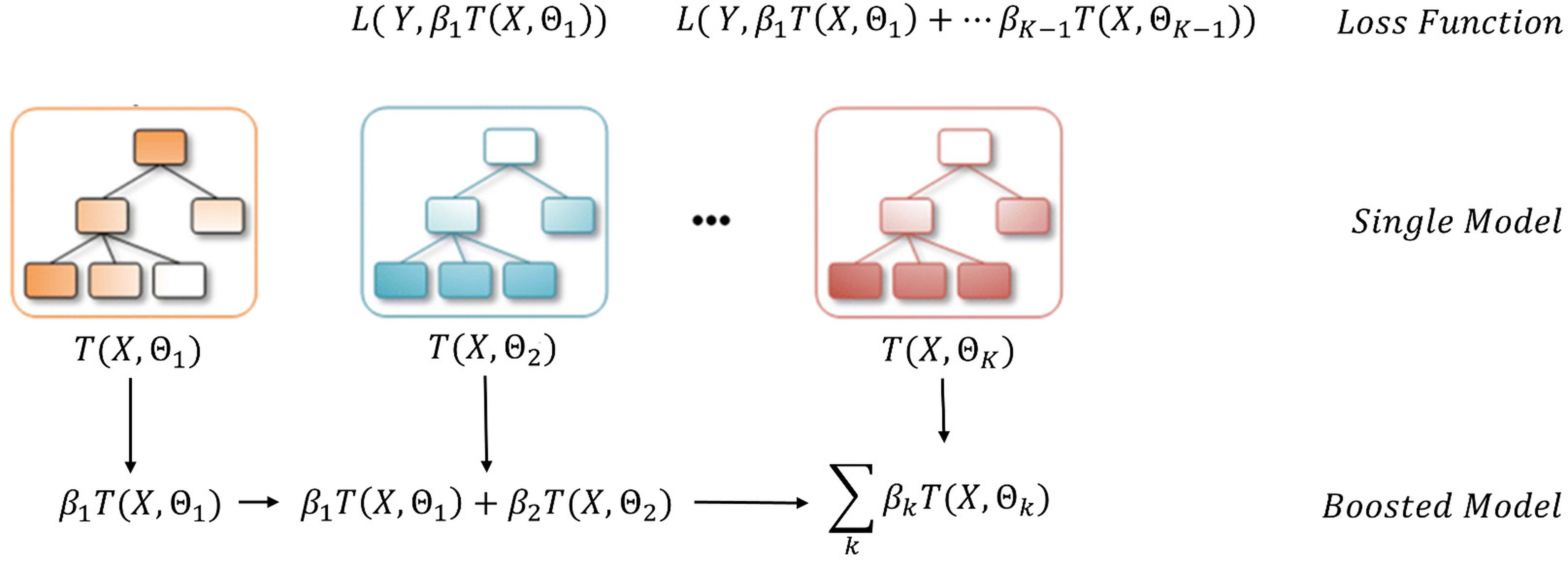}
\captionsetup{format=hang}
\caption{Gradient Boosting Tree model. 
\vspace{0.5em} \\
$T(X,\Theta_k)$ is the best Tree built at step $k$, its parameters $\Theta_k$ are chosen in order to minimise the Loss Function between the target variable $Y$ and the Boosted model of the previous step. \\
The $\beta_k$ parameter is the weight of the Tree, when added in the Boosted Ensemble, this is also chosen with respect to the Loss Function.}
\label{gbm}
\end{figure}

About the choice of Machine Learning algorithms, we use tree-based Machine Learning models, specifically Gradient Boosting Trees (see Figure \ref{gbm}). They retain the enhanced predictive power of Machine Learning models, while having the additional advantage of requiring almost none pre-processing. Because of their structure, they are able to cope with outliers and extreme values easily. \\
In \citep{visani_explanations_2019} a comparison between Logistic Regression and Gradient Boosting Trees applied to CRM, along with an explanation for the increased predictive power of the latter one.
\medskip

\subsection{Model comparison}

\begin{figure}[h]
\centering
\includegraphics[width=0.5\textwidth]{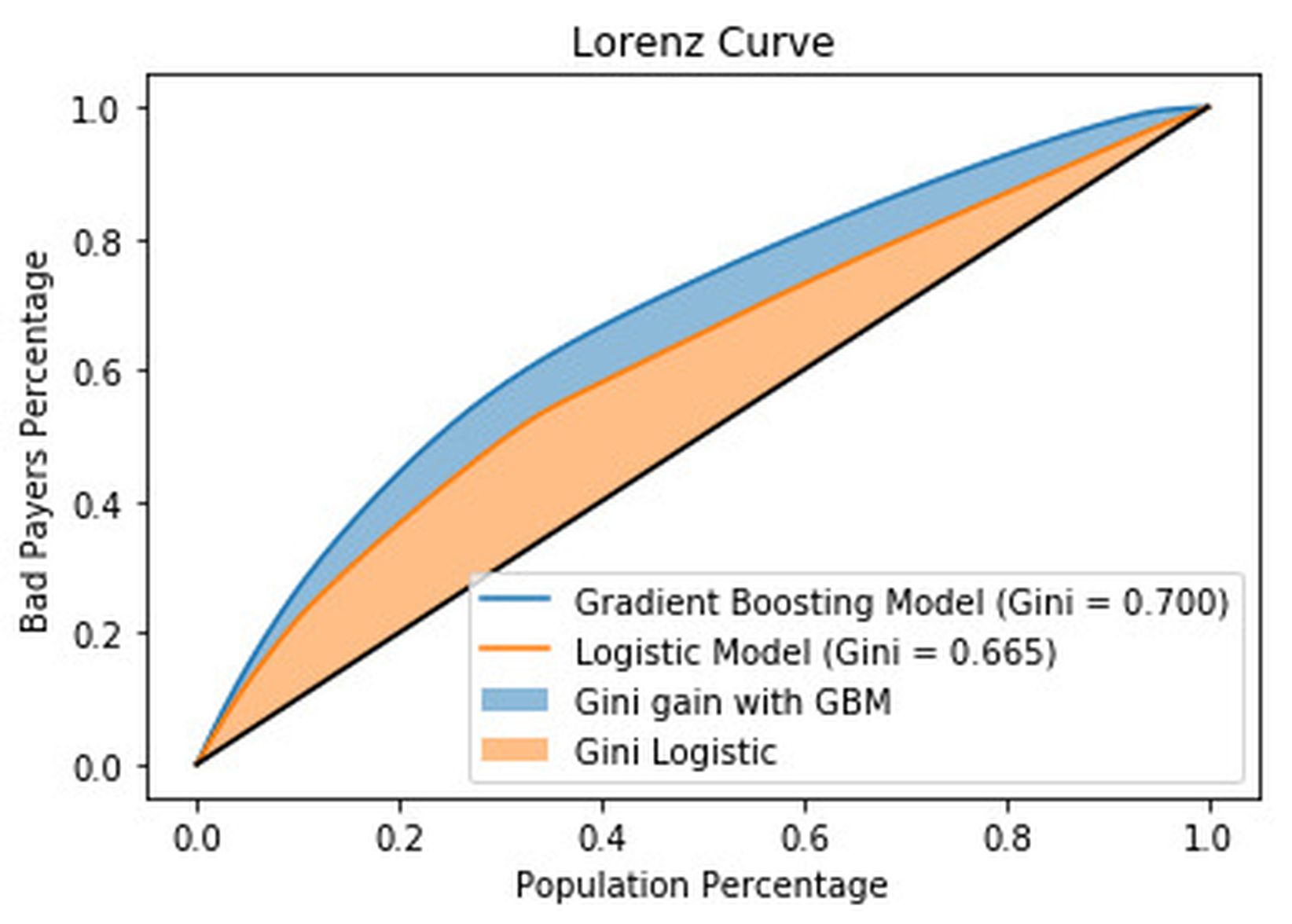}
\captionsetup{format=hang}
\caption{Lorenz Curve comparison: Gradient Boosting vs Logistic Regression}
\label{lorenz_curve}
\end{figure}

The dataset has been divided into Train and Test set, the former is employed to train the models whereas the latter serves for comparison purposes. 
Gradient Boosting's hyperparameters have been tuned by means of grid search and 10-fold cross validation on the training set. On the contrary, Logistic Regression has no hyperparameters to tune and the fit has been carried out on the entire training set, exploiting the Newton-Rhapson iterative procedure for convergence of the coefficients.\\
Performance comparison across the two models is done by means of the Gini Index on the Test set, considered the most reliable figure of merit of the model performances in CRM field \citep{hand_modelling_2001}. In Figure \ref{lorenz_curve}, the Lorentz Curve of the two models is displayed, along with the Gini values.
\medskip

Using Gradient Boosting instead of Logistic Regression, it is possible to recognise an improvement in performance, testified by a Gini increase of more than 3 points. This result is consistent with the recent comparison between Statistical and Machine Learning models applied to the CRM field, carried out by \citep{moscatelli_corporate_2019} at Banca d'Italia.

\subsection{LIME applied to Credit Risk models}

We test LIME on several data points, with the purpose of understanding the logic hidden into the Gradient Boosting model employed. In Table \ref{tab:2}, we report LIME explanations for a “good” user, which has been correctly predicted by the GBM model. Different LIME settings are employed, to demonstrate how a wrong choice of the parameters may yield inconsistent explanations and how the indices are able to spot the instability.
\par

To calculate the indices, LIME is applied 10 times, however the available implementation allows to set the desired number of repetitions. Only the 7 most important features are considered in the explanation.
\medskip


On the left, consistent explanations are achieved, testified by high stability values, whereas on the right a bad choice of the kernel width and the Ridge penalty values bring instability. For more details about the proper choice of such parameters we postpone the reader to \cite{visani_optilime_2020-1}.
\medskip

It is worth noticing LIME results for the stable explanation make sense from an economic and financial standpoint: the key regressors are the Credit Bureau Score (CBS), namely a comprehensive value developed using information provided by the Italian Credit Bureau, and the number of months where unpaid instalments occurred, within the last year.
The user exhibits 0 months with unpaid instalments and falls inside a good class of CBS index.
Such circumstances are the major ones leading Gradient Boosting model to classify him as a good payer.
\par

On the contrary, the unstable LIME method produces different regression lines for each call, making it very hard to trust them since for the same individual we end up with totally different explanations.
\medskip

On a 4 Intel-i7 CPUs 2.90GHz laptop, the indices took 10.23 and 11.54 seconds to be calculated for the left and right settings of the Table \ref{tab:2} respectively.

\begin{table}
\tbl{LIME applied to Gradient Boosting model.
\vspace{0.5em} \\
The sum of the bars' values, along with the intercept, produces the Local Ridge model prediction (denoted as LIME Prediction). The bars' length highlight the specific contribution of each variable: the green ones push the model towards "good payer" prediction, whereas the red ones to "bad payer".
\medskip}
{\begin{tabular}{|c|c|}
\hline
Unit Number: 35  &Unit Number: 35\\ 


Kernel Width: 3  &Kernel Width: 1.3 \\

Ridge Penalty: 1  &Ridge Penalty: 0.001 \\

VSI:  89.44\%  &VSI: 14.17\% \\

CSI: 92.7\%  &CSI: 57.46\% \\ \hline
& \\
\centering
 \includegraphics[width=0.45\linewidth]{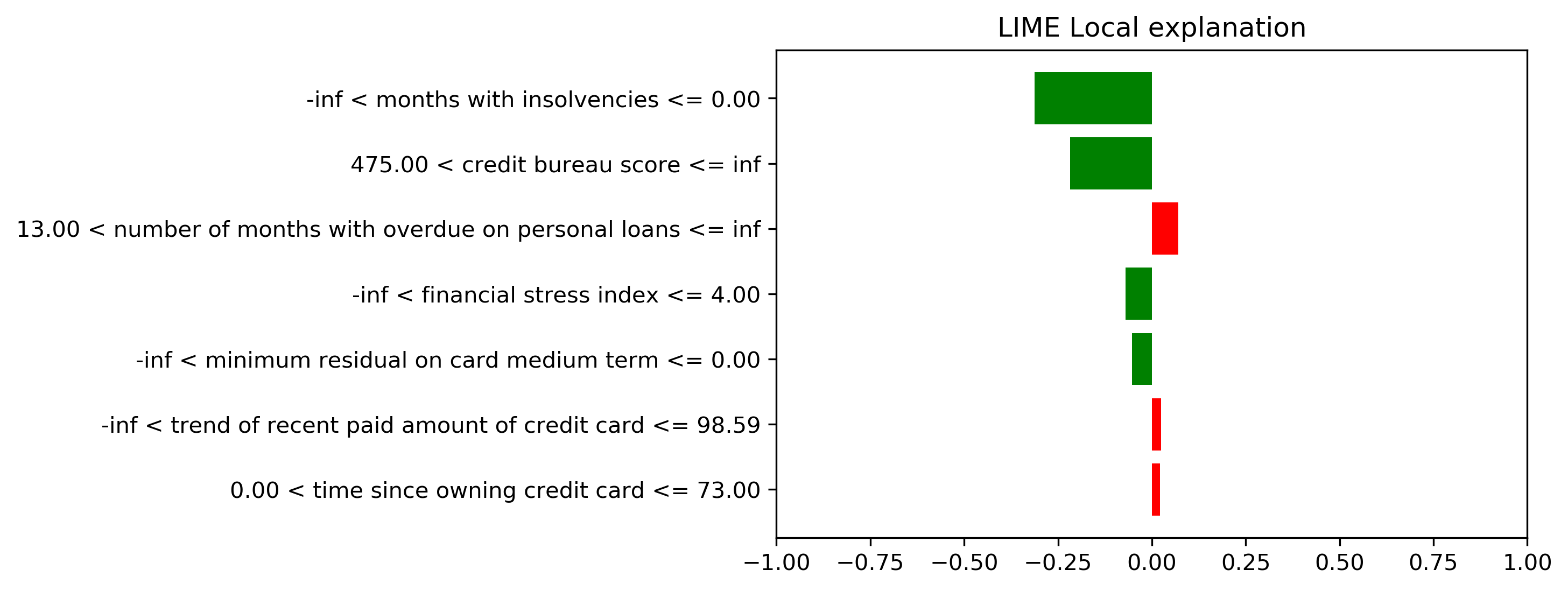}& \includegraphics[width=0.45\linewidth]{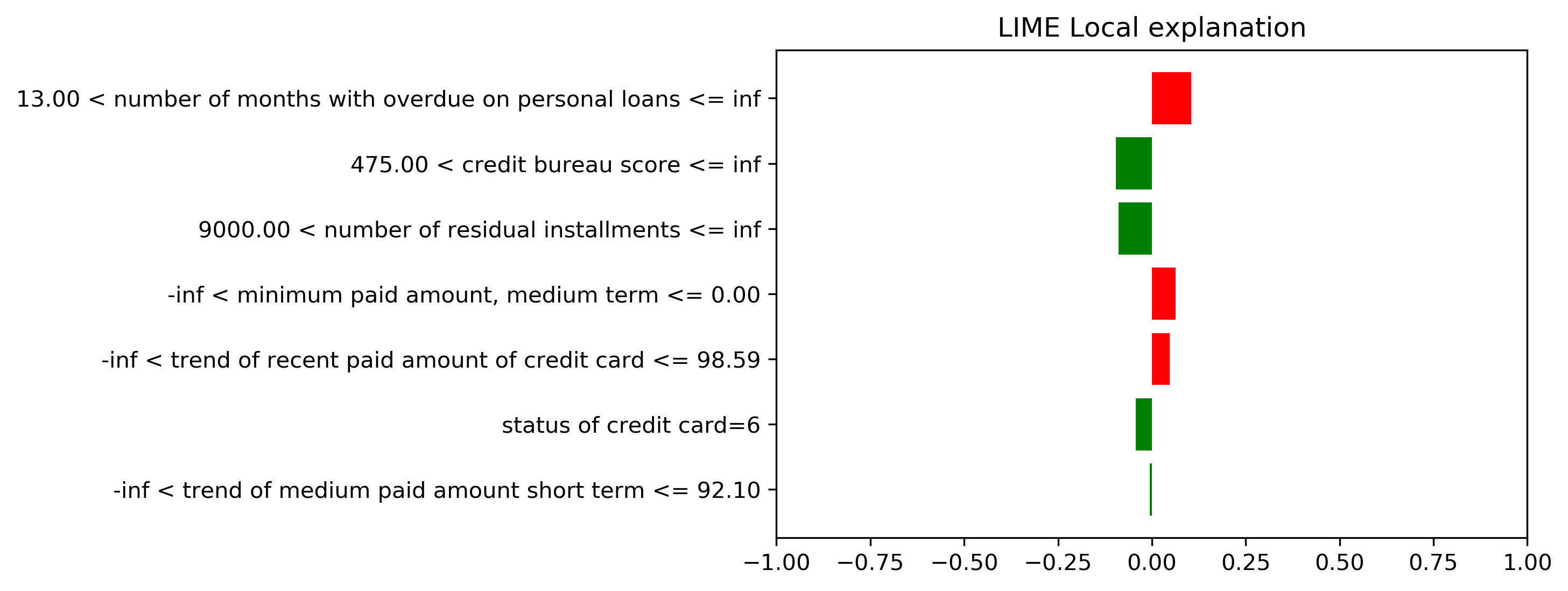} \\ \hline
& \\
 \includegraphics[width=0.45\linewidth]{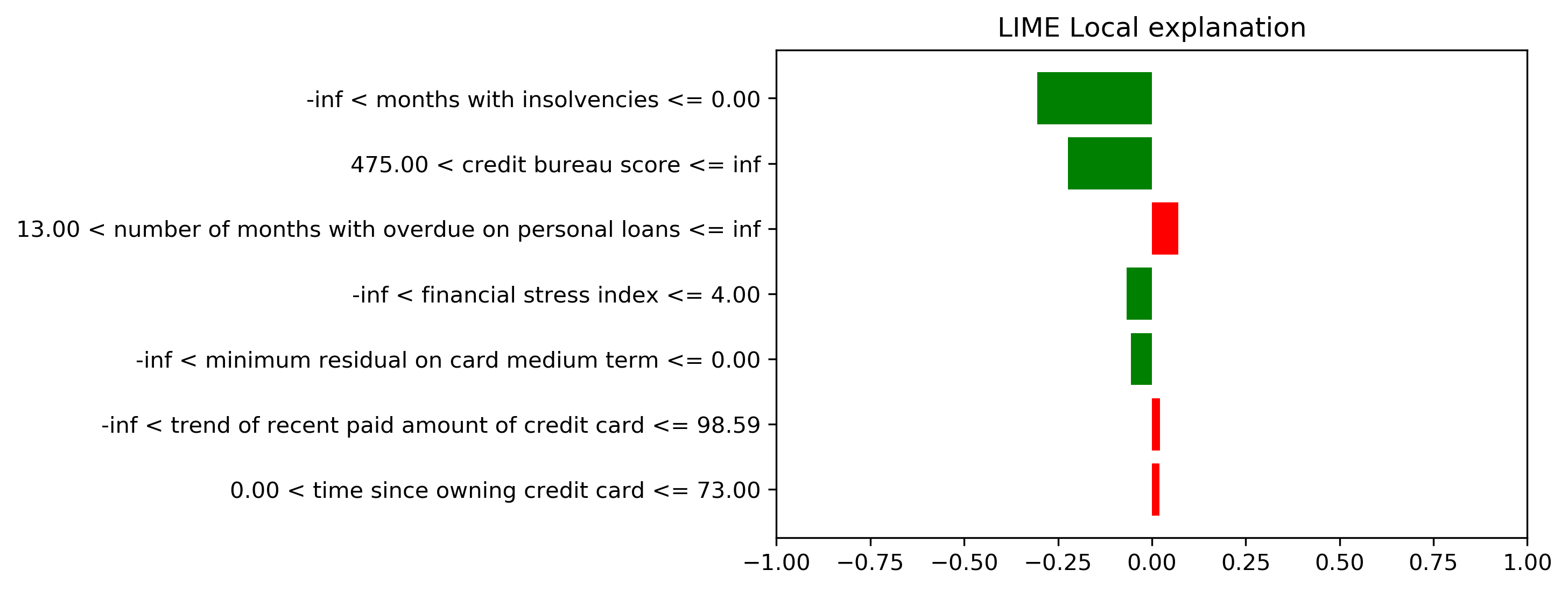}& \includegraphics[width=0.45\linewidth]{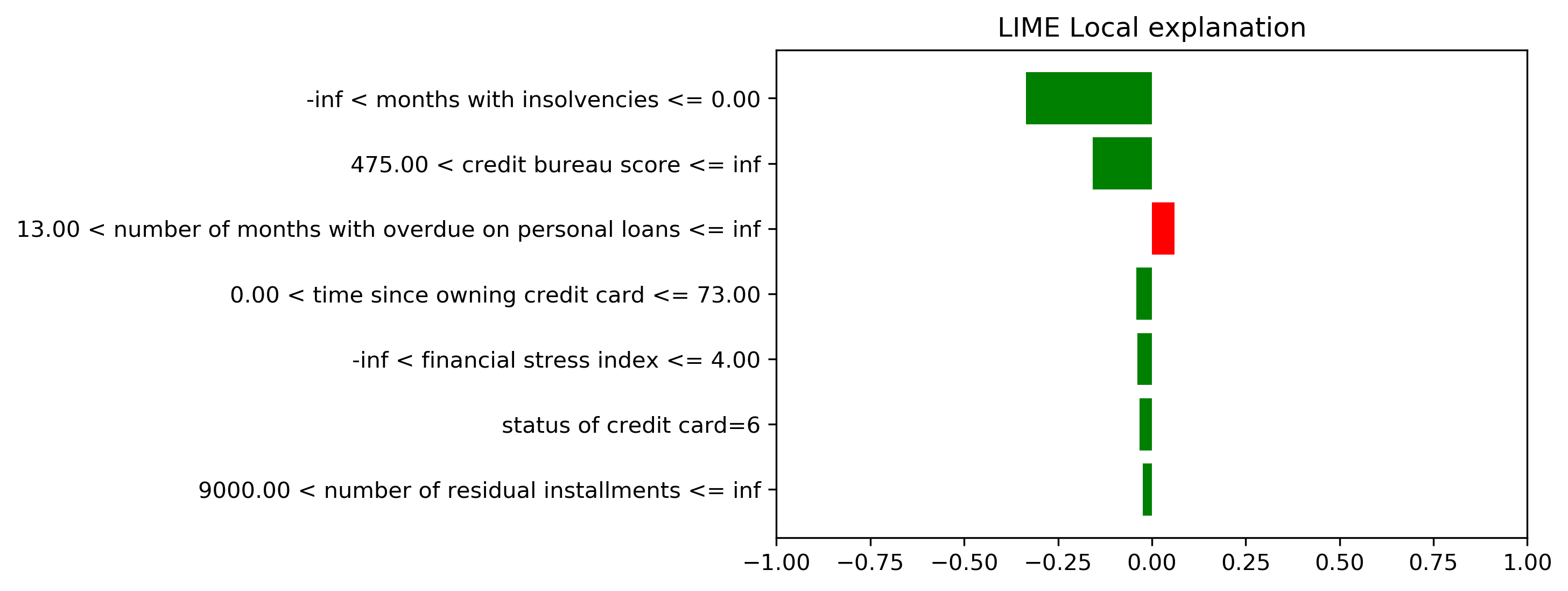} \\
\hline
& \\
 \includegraphics[width=0.45\linewidth]{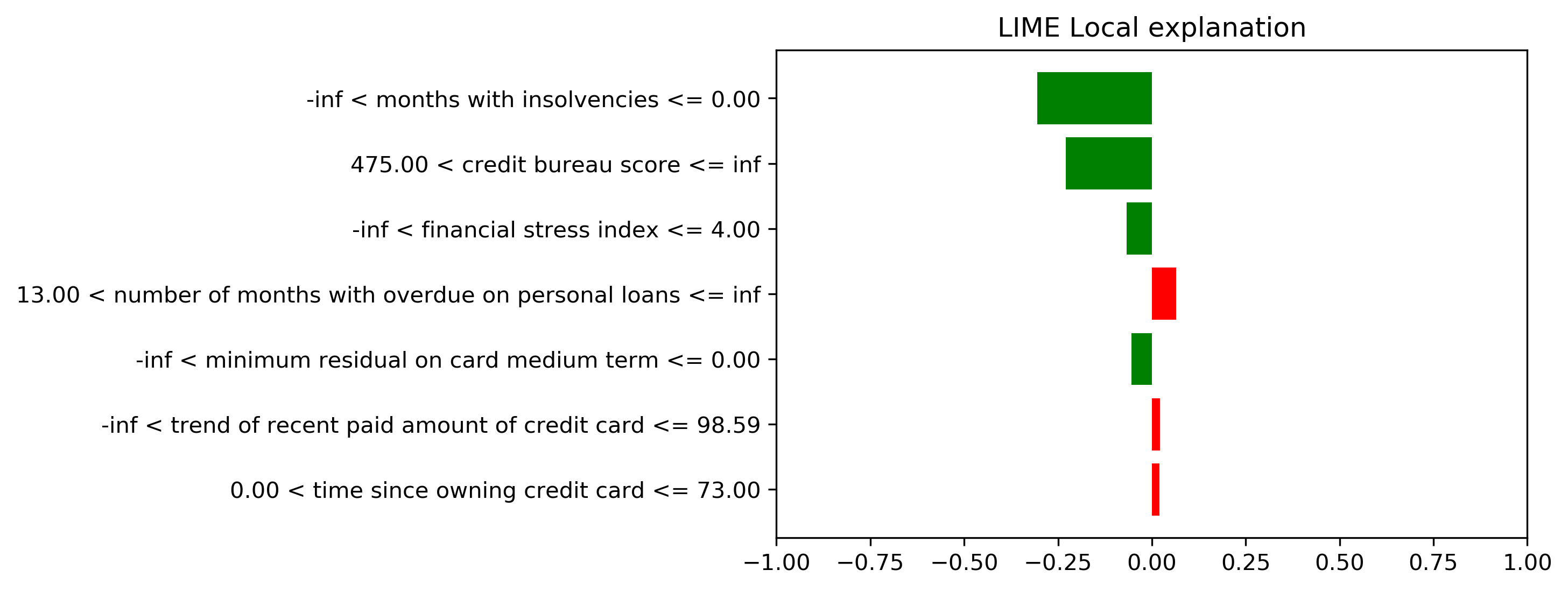}& \includegraphics[width=0.45\linewidth]{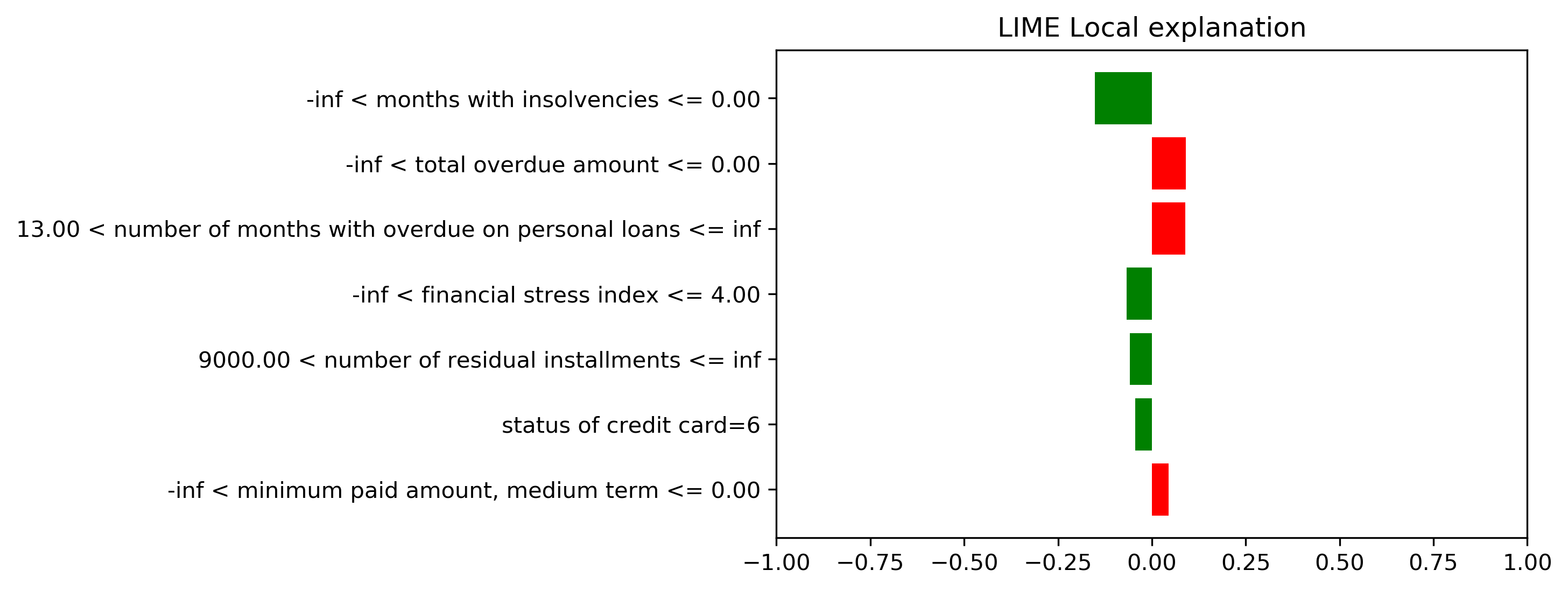} \\ \hline
 \end{tabular}}
 \label{tab:2}
\end{table}

\pagebreak

\section{Discussion and conclusions}

Often Machine Learning models produce more accurate predictions compared to classical models: there was evidence of such trend also in the use case presented above, regarding the CRM field. Credit Risk would therefore benefit from the employment of such more powerful techniques.
\medskip

Regrettably, to date there is no methodology allowing unambiguous explanations of Machine Learning models: in recent years, a number of methods have been proposed, but their consistency and reliability is still a discussion topic.
\medskip

We focus on the LIME technique and apply it successfully to Credit Risk data. Digging further into the method, we try to establish whether LIME is stable, namely if repeated calls of the method, on the same individual, result in very close explanations.
\medskip

To this end, we derive the distribution of the local model coefficients, used by LIME under the hood. Building on top of the coefficients' distribution, we create a stability index, which evaluates whether the coefficients for the same variable among different LIME calls are similar (\CSI). Meanwhile, we monitor whether the variables returned by different LIME calls are the same. This is done using another index: \VSI.
The two complementary indices both range from 0 to 100, where higher values correspond to an higher degree of stability. \\
An application to tabular data is shown in the paper, although the stability framework can be applied also to images and text data, as long as LIME local model is chosen to be Ridge Regression. In fact, the indices formulation relies on the Ridge model properties.
\medskip

When used together, they provide useful insights to the practitioner about the consistency of the trained LIME method: they help understand whether LIME is likely to modify its output at the next call.

We consider it an important step: it improves the trust in LIME as a reliable explanation method and it goes towards meeting the regulator's requests in the CRM field.
\medskip

However, such result just ensures LIME is concordant among different applications: the model may still return explanations not really close to the Machine Learning model. More research still needs to be done in this direction.

\section*{Acknowledgements}
We would like to thank professor Giuliano Galimberti who provided insight and expertise that greatly assisted the research, although he may not agree with all of the interpretations/conclusions of this paper.

\section*{Funding}
We acknowledge financial support by CRIF S.p.A. and Università degli Studi di Bologna.

\bibliographystyle{apacite}
\bibliography{Articolo_CRC}

\end{document}